\documentclass[10pt,twocolumn,letterpaper]{article}

\usepackage[pagenumbers]{cvpr} 
\usepackage{fontawesome5}
\usepackage{fancyhdr}
\usepackage{marvosym}
\definecolor{cvprblue}{rgb}{0.21,0.49,0.74}
\usepackage[pagebackref,breaklinks,colorlinks,allcolors=cvprblue]{hyperref}

\def\paperID{4779} 
\def\confName{CVPR}
\def\confYear{2025}
\usepackage{multirow}
\usepackage{pifont}

\title{Free-viewpoint Human Animation with Pose-correlated Reference Selection}

\author{Fa-Ting Hong$^{1,2}$, Zhan Xu$^{2,}${\textsuperscript{\Letter}}, Haiyang Liu$^2$, Qinjie Lin$^3$, Luchuan Song$^2$, \\ Zhixin Shu$^2$, Yang Zhou$^2$, Duygu Ceylan$^2$, Dan Xu$^{1,}${\textsuperscript{\Letter}}\\
\vspace{-10pt}
\and
$^1$HKUST \quad $^2$Adobe Research \quad $^3$Northwestern University, USA\\
{\tt\small fhongac@connect.ust.hk, \{zhaxu, zshu, yazhou, ceylan\}@adobe.com, haiyangliu1997@gmail.com}\\ {\tt\small qinjielin2018@u.northwestern.edu, lsong11@ur.rochester.edu, danxu@ust.hk}
}

\begin{document}
\twocolumn[{%
\renewcommand\twocolumn[1][]{#1}%
\maketitle

\begin{center}
    \centering
    \captionsetup{type=figure}
    \includegraphics[width=0.99\textwidth]{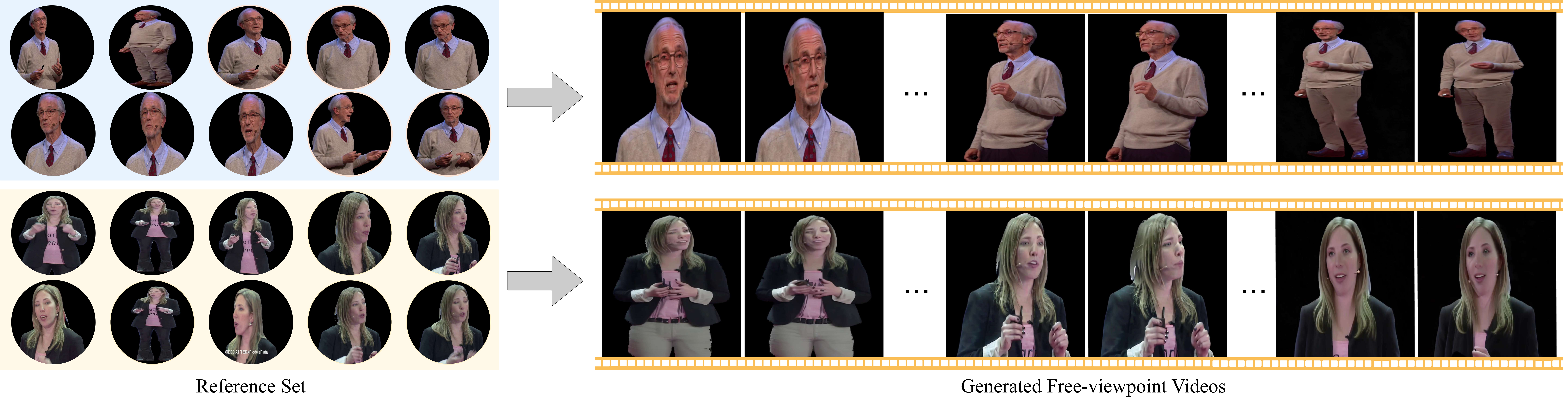}
    \vspace{-8pt}
    \captionof{figure}{In this work, we aim to address the challenging task of \textbf{free-viewpoint human animation synthesis under large viewpoint and camera distance changes}. Our proposed method successfully generates novel-view videos with consistent character appearance across substantial viewpoint shifts.
    }
    \label{fig:teaser}
\end{center}%
}]

 \begin{abstract}

Diffusion-based human animation aims to animate a human character based on a source human image as well as driving signals such as a sequence of poses. Leveraging the generative capacity of diffusion model, existing approaches are able to generate high-fidelity poses, but struggle with significant viewpoint changes, especially in zoom-in/zoom-out scenarios where camera-character distance varies. This limits the applications such as cinematic shot type plan or camera control. We propose a pose-correlated reference selection diffusion network, supporting substantial viewpoint variations in human animation. Our key idea is to enable the network to utilize multiple reference images as input, since significant viewpoint changes often lead to missing appearance details on the human body. To eliminate the computational cost, we first introduce a novel pose correlation module to compute similarities between non-aligned target and source poses, and then propose an adaptive reference selection strategy, utilizing the attention map to identify key regions for animation generation. 
To train our model, we curated a large dataset from public TED talks featuring varied shots of the same character, helping the model learn synthesis for different perspectives. Our experimental results show that with the same number of reference images, our model performs favorably compared to the current SOTA methods under large viewpoint change. We further show that the adaptive reference selection is able to choose the most relevant reference regions to generate humans under free viewpoints. The project website can be found at \href{https://harlanhong.github.io/publications/fvhuman/index.html}{HERE}.
\end{abstract}

\vspace{-10pt}
\section{Introduction}
\vspace{-3pt}
\label{sec:intro}

Reference-based human animation involves animating a given character based on a driving motion sequence. Typically, a single reference image provides the character’s appearance, while the motion sequence is represented as a sequence of skeletal poses. 
Recent advances in reference-based human animation~\cite{hu2024animate, zhu2024champ, hong2024dreamhead, xu2024magicanimate, zhuzero, shao2024human4dit} have achieved significant improvements in terms of both quality and robustness, largely due to the use of diffusion models~\cite{ho2020denoising,song2020denoising}. However, they primarily focus on animating the character from the same viewpoint as the reference image. Although some methods can synthesize turning motions, they assume a fixed character-camera distance, limiting the animation to the same camera shot type as the reference image.

In this work, we address the aforementioned limitation by tackling a more general and challenging scenario: human video generation under dramatic novel viewpoints. 
For instance, given a full-body reference image, our aim to synthesize a close-up video shot of the character’s upper body, or vice versa (see Fig.\ref{fig:teaser}). With such capability, reference-based human animation becomes truly \emph{viewpoint-free}: dedicated camera control is enabled, making it possible to generate multi-shot human videos with high aesthetics.

Dramatic viewpoint changes reveal the limitations of using a single reference image. First, a single image captures only one viewpoint of the subject, limiting the available visual information for generating new viewpoints. For instance, a close-up reference image provides only a portrait, lacking the lower body information required to synthesize a full-body view, while a distant full-body reference image lacks the fine-grain facial details required for generating a high-resolution close-up face image. Additionally, a single-view reference often includes self-occlusions, thus large viewpoint and pose changes may reveal hidden areas of the body not visible in the reference image. These challenges make the synthesis process heavily dependent on the generation capability of the diffusion model. While diffusion models have shown promise in general text-to-video generation, generating realistic human images and videos remains a challenging task.

Intuitively, incorporating multiple reference views provides more comprehensive visual information. 
Our approach leverages this insight by utilizing multiple reference images as input, enabling the extraction of rich visual details from diverse perspectives. However, directly increasing the number of reference images leads to a linear increase in computational complexity. To mitigate this, we employ a selective strategy: rather than considering all spatial locations in the reference feature maps, we select a fixed number of relevant intermediate reference features during generation. This selection strategy allows us to utilize more reference images without excessive computational costs, thereby enhancing the realism of the generated video. Furthermore, since dramatic viewpoint changes reduce alignment between target and reference poses, we introduce an effective attention mechanism based on correlations between target and reference poses. This attention map emphasizes critical regions within the intermediate reference feature map, guiding the model to focus on essential visual cues from the reference images.

We train our diffusion model with the proposed pose correlation mechanism and adaptive reference selection on two large-scale datasets. First, we introduce a new multi-shot TED video dataset ({MSTed}), containing 1,084 unique identities captured from diverse viewpoints and shot types. This comprehensive dataset serves as a valuable resource for training and evaluating models on challenging tasks with substantial viewpoint variation. Second, we use the public DyMVHumans dataset~\cite{zheng2024PKU-DyMVHumans}, which comprises $33$ identities from more than $45$ viewpoints, further validating our method's versatility across varied scenarios. Our evaluations on these datasets demonstrate the effectiveness of our approach. Even with a single reference image, training on such challenging data combined with our architectural improvements achieves performance that surpasses existing methods. Furthermore, our experimental results indicate that incorporating multiple reference images substantially improves the quality of the generated videos, particularly in managing dramatic viewpoint shifts that are challenging for current techniques. Ablation studies further confirm that our reference selection strategy successfully identifies and utilizes critical regions within reference images. 

\begin{figure*}[!t]
  \centering    \includegraphics[width=1\linewidth]{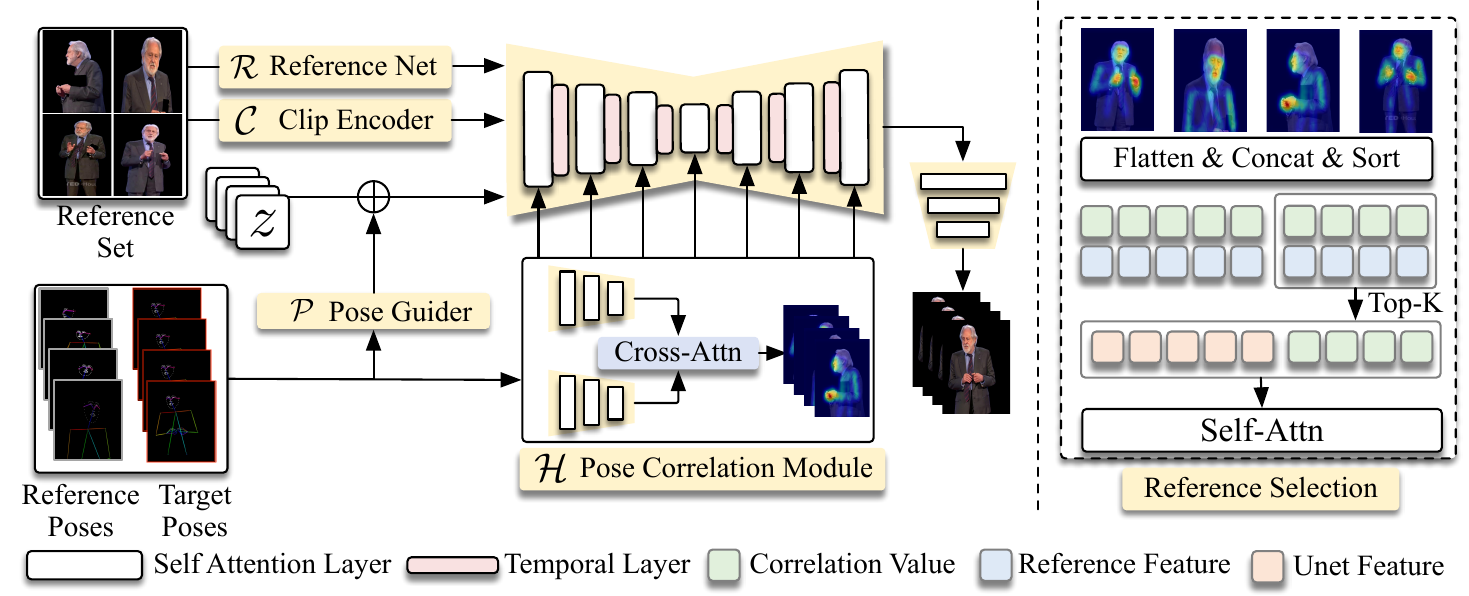}
  \vspace{-20pt}
    \caption{
    The illustration of our framework. Our framework feed a reference set $\{\mathbf{I}_{ref}^i\}_{i=1}^N$ into reference Unet ($\mathcal{R}$) to extract the reference feature. To filter out the redundant information in reference features set, we propose a pose correlation guider to create a correlation map to indicate the informative region of the reference spatially. Moreover, we adopt a reference selection strategy to pick up the informative tokens from the reference feature set according to the correlation map and pass them to the following modules.
    }
    \vspace{-10pt}
    \label{fig:framework}    
\end{figure*}

Our contributions can be summarized as follows:

\begin{itemize}
    \item We propose a novel adaptive reference selection diffusion network for free-viewpoint human animation, enabling flexible human framing and versatile camera shot planning without strict reliance on a specific reference image. 
    
    \item Adaptive reference selection strategy filters redundant features, allowing the use of more reference images without increase of computational cost. This reduces ambiguity caused by missing visual information during significant viewpoint changes. Additionally, our pose correlation module generates attention maps to emphasize critical regions in intermediate reference features, establishing informative correlations between target and reference poses.

    \item We introduce the Multi-Shot TED Video Dataset (MSTed) to advance research in free-viewpoint human video generation. Experimental results on MSTed and DyMVHumans validate the state-of-the-art performance of our approach in generating human videos in novel viewpoints.
    
\end{itemize}

\vspace{-3pt}
\section{Related Works}
\vspace{-3pt}
\label{sec:related}

\noindent\textbf{Human Video Animation.} 
GAN-based models~\cite{siarohin2019first,zhao2022thin,tao2022structure,tao2024learning, hong2022depth, hong2023dagan++,hong2023implicit} utilize the inherent generative ability of adversarial networks~\cite{goodfellow2014generative} to create a spatial transformation for warping reference image to specific pose. FOMM~\cite{siarohin2019first} and DaGAN~\cite{hong2022depth} detect the keypoint of the human image and utilize Taylor approximation approach to estimate the motion flow between between keypoitns pairs. Based on this, \citep{tao2024learning} learn a structure correlation volume from reference image and target image to iteratly refine the motion flow learned from the Taylor approximation. Besides, MCNet~\cite{hong2023implicit} attempt to learn a memory to store the general knowledge of human structure to refine the intermediate features during generation process.
Diffusion-based models~\cite{chen2023motion, xu2024magicanimate, hu2024animate, karras2023dreampose, wang2024disco, zhang2024mimicmotion} have recently demonstrated superior quality and stable controllability in human video generation tasks. These methods typically use a single reference image along with driving signals like SMPL~\cite{loper2023smpl} or DWpose~\cite{yang2023effective} to produce target videos. For example, AnimateAnyone\cite{hu2024animate} employs a double-UNet architecture with a UNet-based ReferenceNet to extract features from the reference image. Champ\cite{zhu2024champ} enhances performance by incorporating multiple driving signals—including semantic maps, depth maps, skeletons, and normal images—into the double-UNet. Human4Dit~\cite{shao2024human4dit} takes a different approach by using camera parameters within a 4D transformer diffusion model to control the generated view.

However, existing approaches are constrained to animating humans under a fixed camera, struggling when there are significant viewpoint variations, such as zoom-in and zoom-out. In this work, we propose a novel approach to address such challenge. Our approach allows for incorporating multiple reference images, providing richer visual information.

\noindent\textbf{Novel View Synthesis for Human} In human-centric novel view synthesis, many existing approaches \cite{hu2024surmo, peng2021neural, liu2019liquid, weng_humannerf_2022_cvpr, wu2023aniportraitgan, kocabas2024hugs} use 3D representations such as NeRF \cite{mildenhall2021nerf} and Gaussian splatting \cite{wu2024recent}.
Specifically, HUGS~\cite{kocabas2024hugs} uses 3D Gaussian Splatting (3DGS) to automatically disentangle and animate a human avatar and static scene from a monocular video with 50-100 frames, leveraging the SMPL body model and allowing deviations to capture details like clothing and hair. HumanNerf~\cite{weng_humannerf_2022_cvpr} optimizes a volumetric representation of a person in a canonical T-pose, paired with a motion field that maps this representation to each frame of the video via backward warps, with the motion field decomposed into skeletal rigid and non-rigid motions generated by deep networks.
AniPortraitGAN \cite{wu2023aniportraitgan} employs 3D GAN techniques to generate portrait images with controllable facial expressions, head poses, and shoulder movements. While these methods facilitate the rendering of human avatars from novel viewpoints, their dependence on 3D representations and human template presents challenges in achieving harmonized, photorealistic appearance and realistic motion. To achieve optimal quality, these methods often rely on multi-view, high-resolution captures of humans in various poses, which limits their generalizability. 
While not specifically designed for human subjects, recent works such as \cite{watson2022novel, chan2023generative, kwak2024vivid, ren2023dreamgaussian4d} employ diffusion models for novel view synthesis; however, they are limited to scenes and objects without motion controllability. In this work, we leverage the generative capabilities of diffusion models to enable novel view synthesis within the challenging setting of human animation. Our approach utilizes a guided motion sequence that encodes the spatial relationship between the camera and the subject. By incorporating multiple reference images, we aim to improve identity preservation and, at the same time, harness the generative potential of diffusion models to infer missing visual content when necessary.

\vspace{-3pt}
\section{Approach}
\vspace{-3pt}
In this work, we address the challenging task of free-viewpoint  human animation under dramatic changes in viewpoint and camera distance. Our method leverages multiple reference images to capture sufficient visual information and employs a selection strategy to exclude irrelevant features, enabling more accurate and realistic video generation. By prioritizing the most informative regions of the reference images, our approach effectively manages significant variations in viewpoint and distance, improving both synthesis quality and efficiency. 
\vspace{-3pt}
\subsection{Overview}
\vspace{-3pt}
In this work, we utilize the popular double UNet architecture~\cite{zhu2024champ, hu2024animate} as our backbone. In addition to the VAE and CLIP encoder, the double UNet structure consists of three key components: the denoising UNet $\mathcal{D}$, the reference UNet $\mathcal{R}$, and the pose guider $\mathcal{P}$. 
The denoising UNet $\mathcal{D}$ is responsible for processing noisy input data and generating high-quality, denoised outputs. The reference UNet $\mathcal{R}$ extracts features from the reference images and aggregates them into the denoising UNet. $\mathcal{P}$ encodes the target poses and feeds them into the denoising UNet as driving conditions.

As illustrated in Figure~\ref{fig:framework}, given a reference set comprising multiple reference images $\{\mathbf{I}_{\text{ref}}^i\}_{i=1}^N$ and their corresponding pose images $\{\mathbf{P}_{\text{ref}}^i\}_{i=1}^N$ ($N$ is the number of reference number), our goal is to synthesize novel-view videos guided by a target pose sequence $\{\mathbf{P}_{\text{tgt}}^j\}_{j=1}^T$ ($T$ is the length of driven pose sequence). Our method first processed all reference images through the reference UNet $\mathcal{R}$ and CLIP encoder $\mathcal{C}$ to extract spatial reference features and semantic CLIP features, respectively. These features are subsequently fed into the denoising UNet for video generation. To identify the informative regions of each reference image contributing to the target frames, a pose correlation module $\mathcal{H}$ is proposed to compute correlation maps between the reference and target poses, highlighting the most relevant regions. Based on the correlation maps, a reference selection strategy is employed to filter and forward only the informative reference features to the subsequent network stages. Detailed descriptions of each component are provided in the following sections.

\begin{figure}[t]
  \centering
    \includegraphics[width=1\linewidth]{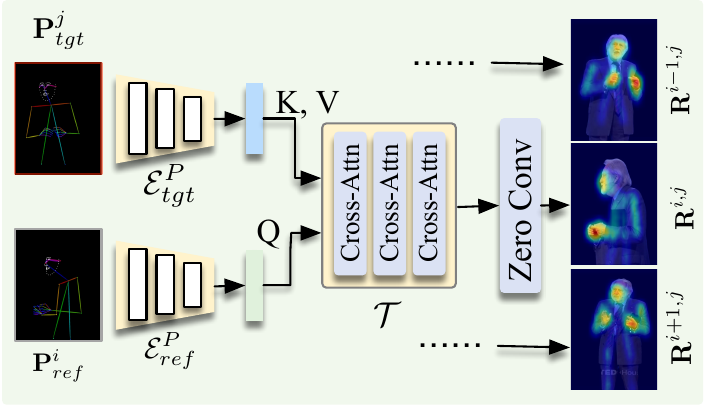}
    \vspace{-15pt}
    \caption{The illustration of pose correlation module (\textbf{PCM}). Each reference pose $\mathbf{P}_{ref}^i$ will be fed into PCM with each target pose $\mathbf{P}_{tgt}^j$ to compute a correlation map, which indicate the informative region of input reference image.
    }
    \vspace{-10pt}
    \label{fig:correlation}    
\end{figure}
\vspace{-3pt}
\subsection{Multiple Reference Learning}
\vspace{-3pt}

Previous work, such as~\cite{hu2024animate,xu2024magicanimate}, utilized a single reference image to synthesize videos based on given pose signals. However, this approach faces substantial challenges when handling dramatic viewpoint changes, as a single reference image often lacks sufficient visual content to cover all necessary appearance details. To overcome this limitation, we propose to leverage multiple reference images to aggregate a richer set of visual information, enhancing the quality and accuracy of the generation process.

As illustrated in Figure~\ref{fig:framework}, we utilize multiple reference images $\{\mathbf{I}_{\text{ref}}^i\}_{i=1}^N$ from diverse viewpoints. This enables the reference images to provide comprehensive coverage of the human’s appearance across various perspectives. At test time, given multiple reference images as input, we first use reference UNet $\mathcal{R}$ to extract the reference feature: $ \{\mathbf{F}^i_{l}\}_{l=1}^L = \mathcal{R}(\mathbf{I}_{\text{ref}}^i)$, 
where $L$ denotes the number of layers in the reference UNet. Then our network selects the most relevant regions from 
$\{\mathbf{F}^i_{l}\}_{l=1}^L$ based on the learned pose correlation as described below, enhancing the generation quality.

\noindent\textbf{Pose Correlation Learning.} Introducing multiple references inevitably adds redundant visual information, which increases computation and distracts the model. To mitigate this, we propose a pose correlation learning module to identify informative regions in the reference images by leveraging correlations between the target pose and the reference pose.

As shown in Figure~\ref{fig:correlation}, given a target pose $\mathbf{P}_{\text{tgt}}^j$ and a reference pose $\mathbf{P}_{\text{ref}}^i$, we first utilize two pose encoders ($\mathcal{E}_{\text{ref}}^P$ and $\mathcal{E}_{\text{tgt}}^P$, with identical structures but do not share weights) to extract the reference and target pose features:
\begin{equation}
    \mathbf{F}^i_{\text{ref}}, \mathbf{F}^j_{\text{tgt}} = \mathcal{E}_{\text{ref}}^P(\mathbf{P}_{\text{ref}}^i), \mathcal{E}_{\text{tgt}}^P(\mathbf{P}_{\text{tgt}}^j).
\end{equation}

To capture the correlation attention map $R^{i,j}$ between the target pose and the reference pose, we employ a transformer block containing a series of cross-attention layers. The target pose feature $\mathbf{F}^j_{\text{tgt}}$ serves as the \textbf{key} and \textbf{value}, while the reference pose feature $\mathbf{F}^i_{\text{ref}}$ acts as the \textbf{query}:
\begin{equation}
    \mathbf{R}^{i,j} = f_{\text{zero}} \circ \mathcal{T}(\mathbf{W_q} \mathbf{F}^i_{\text{ref}}, \mathbf{W_k} \mathbf{F}^j_{\text{tgt}}, \mathbf{W_v} \mathbf{F}^j_{\text{tgt}}),
    \label{eq:mulR}
\end{equation}
where $f_{\text{zero}}$ is a convolutional layer with zero-initialized weights to produce the correlation attention map. $\mathcal{T}$ represents the transformer block, and $\mathbf{W_q}$, $\mathbf{W_k}$, and $\mathbf{W_v}$ are learnable parameters used to project the pose features.

\noindent\textbf{Reference Feature Enhancement.} {The pose correlation module generates a set of correlation maps $\{\mathbf{R}^{i,j}\}_{i=1,j=1}^{N,T}$ for each reference-target pose pair $(\mathbf{P}_{\text{ref}}^i\in \{\mathbf{P}_{\text{ref}}^i\}_{i=1}^N, \mathbf{P}_{\text{tgt}}^j \in \{\mathbf{P}_{\text{tgt}}^j\}_{j=1}^T)$. After that, we employ a simple multiplication} to fuse the $i$-th reference feature ($\mathbf{F}^i_l \in \mathbb{R}^{w_l \times h_l}, \, l = 1, \dots, L$) with the correlation map $\mathbf{R}^{i,j}$ at each layer $l$:
\begin{equation}
    \mathbf{F}_{\text{cor,l}}^{i,j} = f_{\text{Inter}}(\mathbf{R}^{i,j}) \times \mathbf{F}^i_l,
\end{equation}
where $f_{\text{Inter}}$ is a bilinear interpolation function that adjusts the size of $\mathbf{R}^{i,j}$ to match that of $\mathbf{F}^i_l$ at $l$-th layer. 

Subsequently, all highlighted features $\mathbf{F}_{\text{cor},l}^{i,j}, i \in [1, N]$ are concatenated and fed into the denoising UNet $\mathcal{D}$. These features are concatenated with the UNet's intermediate latents by along spatial dimension before fed into each spatial self-attention layer. 
This design allows the model to concentrate on the informative regions of the reference images highlighted by the correlation map, and thus improves the accuracy of novel viewpoint video synthesis.

\vspace{-3pt}
\subsection{Adaptive Reference Selection Strategy}
\vspace{-3pt}

\noindent\textbf{Correlation-Guided Reference Selection.} The above feature manipulation enables the model to focus on highlighted regions of the reference feature map. However, increasing the number of reference images significantly raises the computational burden, particularly during the iterative denoising process in the diffusion model. To address this, we propose a selection strategy that filters and passes only the most important visual features to the denoising UNet.

As shown in Figure~\ref{fig:framework}, to synthesize the $j$-th frame, we compute a set of correlation maps $\{\mathbf{R}^{i,j}\}_{i=1}^N$ from the target pose $\mathbf{P}_{\text{tgt}}^j$ and all reference poses $\{\mathbf{P}_{\text{ref}}^i \in \mathbb{R}^{w \times h}\}_{i=1}^N$. We flatten all correlation maps and concatenate them (assuming that the correlation maps have the same resolution as the reference features after interpolation):
\begin{equation}
\begin{aligned}
    \mathbf{r}_j &= \mathbf{r}_{1,j} || \mathbf{r}_{2,j} || \dots || \mathbf{r}_{N,j}, \\
    \mathbf{f}_l &= \mathbf{f}_l^1 || \mathbf{f}_l^2 || \dots || \mathbf{f}_l^N,
\end{aligned}
\end{equation}
where $\mathbf{r}_{i,j} \in \mathbb{R}^n$, with $n = w_l \times h_l$, represents the flattened correlation map $\mathbf{R}^{i,j}$. The symbol ``$||$'' denotes the concatenation operator. Similarly, we obtain the flattened reference features $\mathbf{f}_l$ from reference features $\{\mathbf{F}_l^i\}_{i=1}^N$.

We select the most informative reference features from $\mathbf{f}_l$ based on the correlation values $\mathbf{r}_j$. Specifically, we sort the correlation values $\mathbf{r}_j$ and select the top $K_l$ values at $l$-th layer in denoising UNet as follows:
\begin{equation}
\begin{aligned}
    \mathbf{f}_l^{K_l} &= \{f_l^i \mid i \in \text{argsort}(\mathbf{r}_j)[:K_l]\}, \\
    \mathbf{r}_j^{K_l} &= \{r_j^i \mid i \in \text{argsort}(\mathbf{r}_j)[:K_l]\},
\end{aligned}
\end{equation}
where $\text{argsort}(\mathbf{r}_j)$ returns the indices of $\mathbf{r}_j$ sorted in descending order.

Using this approach, we fuse the top $K_l$ reference features with their corresponding correlation values:
\begin{equation}
    \mathbf{f}_{\text{cor},l}^j = \mathbf{f}_l^{K_l} \times \mathbf{r}_j^{K_l}.
\end{equation}

Consequently, the selected correlated features $\mathbf{f}_{\text{cor},l}^j$ are aggregated with the denoising UNet features and passed to subsequent structures, such as the spatial self-attention layer and temporal layer. 

\noindent\textbf{Compensated Reference Feature Sampling.}  
Relying strictly on top-K selection during training can lead the network to a local minimum, as the argsort operation is non-differentiable. To encourage the learning process towards a globally optimal solution, we sample additional $K_l$ reference features uniformly from all reference features:
\begin{equation}
    \mathbf{f}_{\text{random},l}^j = \mathcal{S}_{uni}(\mathbf{f}_l,K_l),
\end{equation}

$\mathcal{S}_{uni}(\cdot, K_l)$ represents uniformly sampling $K_l$ elements. This allows the model to better capture the overall context. 
By introducing uniform sampling, the correlation map $\{\mathbf{R}^{i,j}\}_{i=1,j=1}^{N,T}$ and the pose correlation module can be trained with scattered gradient more stably.

During training, both $\mathbf{f}_{\text{cor},l}^j$ and $\mathbf{f}_{\text{random},l}^j$ are injected into the denoising UNet. During testing, only $\mathbf{f}_{\text{cor},l}^j$ is used, which encourages the model focus on correlated regions and reduces the memory footprint.

\vspace{-3pt}
\subsection{Network Learning}
\vspace{-3pt}
\noindent\textbf{Training Phase.} To simulate the randomness of reference images during testing, we use randomly selected reference images in each training step. The number of reference images is randomly sampled between $1$ and $M$ (during testing, more than $M$ reference images can be used). Similar to previous double UNet methods~\cite{zhu2024champ,hu2024animate}, our training process is divided into two stages, \ie, image training step and temporal traning step. We follow the training manner as in AnimateAnyone~\cite{hu2024animate} and Champ~\cite{zhu2024champ}:

\noindent\textbf{Testing Phase.} During the testing phase, our diffusion model takes multiple reference images and their corresponding poses, along with a target pose sequence, as input. To handle target videos with more than 12 frames, we adopt a temporal aggregation method~\cite{tseng2023edge} to concatenate multiple clips during inference. This approach enables the generation of long-duration video outputs.

\noindent\textbf{Objective Loss.} In this work, we predict the noise $\epsilon_\theta$ at time $t$ and quantify the expected mean squared error (MSE) between the actual noise $\epsilon$ and the predicted noise $\epsilon_\theta$ conditioned on the timestep $t$.
\vspace{-3pt}
\subsection{MSTed Dataset Construction}
\vspace{-3pt}
\begin{table*}[ht]
  \centering
  \resizebox{1\linewidth}{!}{
        \begin{tabular}{lccccccc}
        \toprule
         \multirow{2}*{\textbf{Dataset}} & \multicolumn{5}{c}{\textbf{Dataset Statistics}} & \multicolumn{2}{c}{\textbf{Identity Split}} \\
         \cmidrule(lr){2-6} \cmidrule(lr){7-8}
         & \textbf{Identities} & \textbf{Clips} & \textbf{Total Duration (hrs)} & \textbf{Camera Distance Changes} & \textbf{Viewpoint Changes} & \textbf{Train IDs} & \textbf{Test IDs} \\ 
        \midrule
        DyMVHumans~\cite{zheng2024PKU-DyMVHumans} & 33 & 1,964 & 5.376 & \ding{55}  & \ding{51} & 30 & 3 \\
        MSTed (Ours) & 1,084 & 15,260 & 29.923 & \ding{51} & \ding{51} & 1,000 & 84 \\
        \bottomrule
        \end{tabular}
  }
  \vspace{-10pt}
\caption{Comparison between our proposed MSTed dataset and the existing DyMVHumans dataset~\cite{zheng2024PKU-DyMVHumans}. MSTed offers a significantly larger scale with 1,084 identities and 15,260 clips spanning approximately 30 hours of video. Furthermore, MSTed includes diverse real-world variations in camera distances and viewpoints, making it more representative of in-the-wild scenarios.}
  \vspace{-15pt}
\label{tab:dataset}
\end{table*}

Currently, there are no publicly available video datasets with substantial variations in both viewpoints and camera-to-subject distances. Existing datasets, such as MVHuman~\cite{xiong2024mvhumannet} (only images are provided on the official website) and DyMVHumans~\cite{zheng2024PKU-DyMVHumans}, rely on multi-camera setups to record human videos simultaneously. However, these datasets are captured in controlled studio environments, where the camera-to-subject distance remains uniform across cameras. As a result, they fail to replicate the complexity of real-world scenarios, such as zoom-in and zoom-out effects.

To address this limitation and advance research in this area, we introduce a novel multi-shot TED video dataset (\textbf{MSTed}), designed to capture significant variations in viewpoints and camera distances. TED videos were chosen for their diverse real-world settings, professional quality, rich variations in human presentations, and broad public availability, making them an ideal foundation for a comprehensive and realistic multi-shot video dataset. MSTed dataset comprises 1,084 unique identities and 15,260 video clips, totaling approximately 30 hours of content, as detailed in Table~\ref{tab:dataset}. The dataset was constructed by downloading publicly available TED videos and leveraging DINOv2~\cite{oquab2023dinov2} to segment them into clips based on frame similarity, with each clip corresponding to a single shot. To ensure quality and relevance, a YOLO detector~\cite{redmon2016you} was employed to filter out clips where humans were not consistently present across all frames or where multiple individuals appeared. This rigorous process ensures that MSTed consists exclusively of high-quality, single-person shots, enabling precise human animation tasks while minimizing noise. By structuring the dataset this way, we provide a valuable resource for training and evaluating models that rely on multi-shot video data. We will release the dataset and preprocessing code once accepted.

\vspace{-3pt}
\section{Experiments}
\vspace{-3pt}

In this section, we present quantitative and qualitative experiments to validate the effectiveness of our multiple-reference selection diffusion network. More results and details are reported in \emph{Supplementary Materials}.
\vspace{-3pt}
\subsection{Experimental Settings}
\vspace{-3pt}

\noindent\textbf{Dataset.} In this work, we conduct our experiments on our collected MSTed dataset and the publicly available DyMVHumans dataset~\cite{zheng2024PKU-DyMVHumans}. To ensure consistency and focus on single-person scenarios, we filtered out videos in DyMVHumans that contained multiple individuals. This preprocessing step aligns DyMVHumans with the characteristics of the MSTed dataset, which consists exclusively of high-quality single-person video clips, facilitating a fair comparison and robust evaluation. 

\noindent\textbf{Metrics.} To evaluate our method, we employ several commonly used metrics. We use the $\mathcal{L}_1$ loss to assess pixel-level accuracy, while PSNR, LPIPS, and FVD are used to evaluate video quality. Additionally, we adopt the Mean Opinion-Weighted Indicator for Video Evaluation (MOVIE)~\cite{seshadrinathan2009motion} to measure appearance consistency.

\begin{figure*}[ht]
  \centering
    \includegraphics[width=1\linewidth]{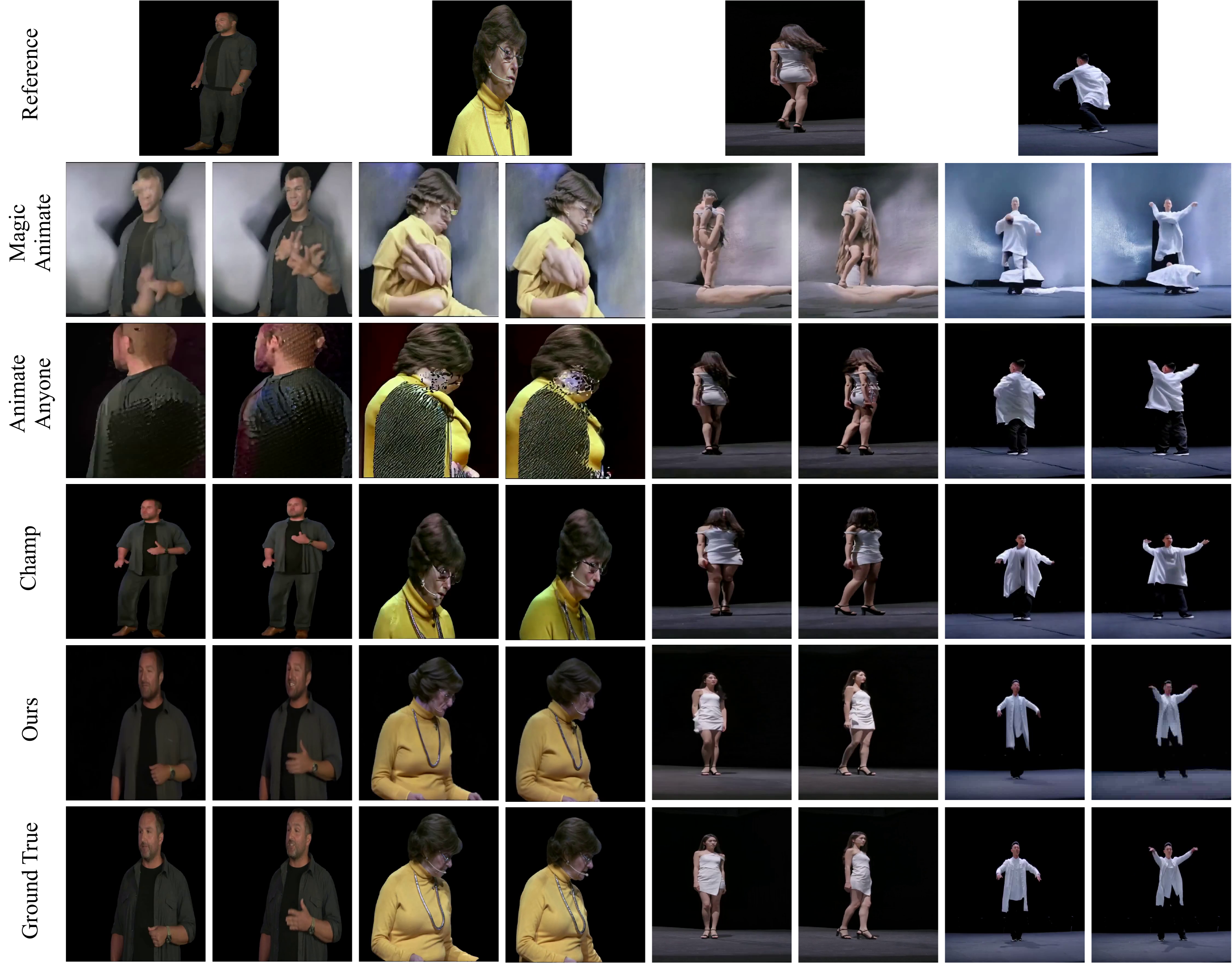}
    \vspace{-15pt}
    \caption{Qualitative results in Multi-Shot Ted dataset and DyMVHuman dataset. Compared with compared methods, our model can achieve high quality while maintaining the appearance consistency.
    }
    \vspace{-10pt}
    \label{fig:cmp}    
\end{figure*}

\vspace{-3pt}
\subsection{Comparison with State-of-the-art Methods}
\vspace{-3pt}

\begin{table}[t]
  \centering
  \resizebox{1\linewidth}{!}{
        \begin{tabular}{lccccc}
        \toprule
        
        Model & $\mathcal{L}_1$ $\downarrow$ & PSNR $\uparrow$ &  LPIPS $\downarrow$ & MOVIE  $\downarrow$ & FVD $\downarrow$  \\
        \midrule
        MagicAnimate~\cite{xu2024magicanimate} &154.02&27.92&0.5984&119.33&35.08 \\
        AnimateAnyone~\cite{hu2024animate} &113.69&29.38&0.5458&94.93&33.10 \\
        Champ~\cite{zhu2024champ} &81.69&30.87&0.4618&67.84&25.68 \\
        Ours(R=1)&\textbf{78.91}&\textbf{32.18}&\textbf{0.2045}&\textbf{56.53}&\textbf{20.88} \\
        Ours(R=2)&\textbf{74.20}&\textbf{32.49}&\textbf{0.1869}&\textbf{55.60}&\textbf{7.044} \\
        \bottomrule
        \end{tabular}
}
    \vspace{-10pt}
\caption{Quantitative results comparision on the MSTed dataset. The results show in table indicate that our method perform better. And the results can be better with the increase of reference number. ``R=2'' means our model uses 2 reference images.} 
\label{tab:comp_ted}
    \vspace{-10pt}
\end{table}

\begin{table}[t]
  \centering
  \resizebox{1\linewidth}{!}{
        \begin{tabular}{lccccc}
        \toprule
        
        Model & $\mathcal{L}_1$ $\downarrow$ & PSNR $\uparrow$ &  LPIPS $\downarrow$ & MOVIE  $\downarrow$ & FVD $\downarrow$ \\
        \midrule
        MagicAnimate~\cite{xu2024magicanimate} &132.64&27.87&0.6583&109.74&51.433 \\
        AnimateAnyone~\cite{hu2024animate} &56.30&33.58&0.2179&42.56&12.300 \\
        Champ~\cite{zhu2024champ} &61.65&30.87& 0.4388&63.21&45.762 \\
        Ours(R=1)&\textbf{56.47}&\textbf{34.22}&\textbf{0.1660}&\textbf{39.70}&\textbf{9.047} \\
        Ours(R=2)&\textbf{50.37}&\textbf{34.78}&\textbf{0.1435}&\textbf{39.21}&\textbf{8.782} \\
        Ours(R=5)&\textbf{49.43}&\textbf{34.84}&\textbf{0.1400}&\textbf{39.10}&\textbf{7.656} \\
        Ours(R=10)&\textbf{35.27}&\textbf{35.02}&\textbf{0.1383}&\textbf{34.35}&\textbf{5.459} \\
        \bottomrule
        \end{tabular}
}
\vspace{-10pt}
\caption{Quantitative results on DyMVHumans dataset~\cite{zheng2024PKU-DyMVHumans}. Our model can accept 10 reference images and get good results. }
\vspace{-10pt}
\label{tab:comp_dyhuman}
\end{table}

In Table~\ref{tab:comp_ted} and Table~\ref{tab:comp_dyhuman}, we compare our method with current state-of-the-art (SOTA) methods (\ie, MagicAnimate~\cite{xu2024magicanimate}, AnimateAnyone~\cite{hu2024animate}, and Champ~\cite{zhu2024champ}) on two datasets.  
Our model is trained in a multiple-reference setting. For a fair comparison, we also perform an experiment where we use only a single reference image as input to match the settings of one-reference approaches. The results show that our model performs favorably compared to other competitive methods under dramatic viewpoint change.  

As shown in Table~\ref{tab:comp_dyhuman}, even when using only one reference image during inference, our model—trained with multiple references—achieves the highest quality. For example, in the DyMVHumans dataset, our method achieves the lowest FVD value ($9.047$) and the highest PSNR value ($34.22$) compared to other methods using a single reference image. This demonstrates that training with multiple reference images enables our model to associate information from different perspectives, even when only one reference is available during inference. Furthermore, compared to one-reference approaches, our method achieves better appearance consistency. For instance, in the MSTed dataset, our model obtains the lowest MOVIE value ($56.53$). This is because other methods struggle when the reference and target poses differ significantly in viewpoint.

Additionally, we visualize some comparison results in Figure~\ref{fig:cmp}. As shown, even when limited to a single reference image, our method achieves the best results. While Champ~\cite{zhu2024champ} leverages multiple modalities as driving signals and achieves strong results, our method synthesizes better appearance-consistent outputs when using only skeleton poses as driving signals. This is attributed to our use of a correlation map, which effectively highlights the most critical regions for the generation process.  

These results verify the effectiveness of our proposed adaptive reference selection diffusion network for free-viewpoint human animation.

\vspace{-3pt}
\subsection{Ablation Study}
\vspace{-3pt}

\begin{table}[t]
  \centering
  \resizebox{1\linewidth}{!}{
        \begin{tabular}{lccccc}
        \toprule
       
        Model & $\mathcal{L}_1$ $\downarrow$ & PSNR $\uparrow$ &  LPIPS $\downarrow$ & MOVIE  $\downarrow$ & FVD $\downarrow$ \\
        \midrule
        baseline &86.77&30.80&0.2377&66.97&26.32\\ 
        baseline+2ref &78.39&31.94&0.2180&59.75&9.82\\ 
        baseline+2ref+$\mathcal{H}$  &\textbf{74.18}&32.20&0.2070&56.77&7.60\\ 
        Ours (R=2)&\textbf74.20&\textbf{32.49}&\textbf{0.1869}&\textbf{55.60}&\textbf{7.04} \\
        \bottomrule
        \end{tabular}
}
\vspace{-10pt}
\caption{Ablation study comparing the performance of our method with various configurations, including the baseline, baseline with two references, and the addition of the pose correlation module.$\mathcal{H}$ means the pose correlation module. Our proposed method with two reference images achieves the best results across all metrics.}
\vspace{-15pt}
\label{tab:abla_ele}
\end{table}

\begin{figure*}[t]
  \centering
    \includegraphics[width=1\linewidth]{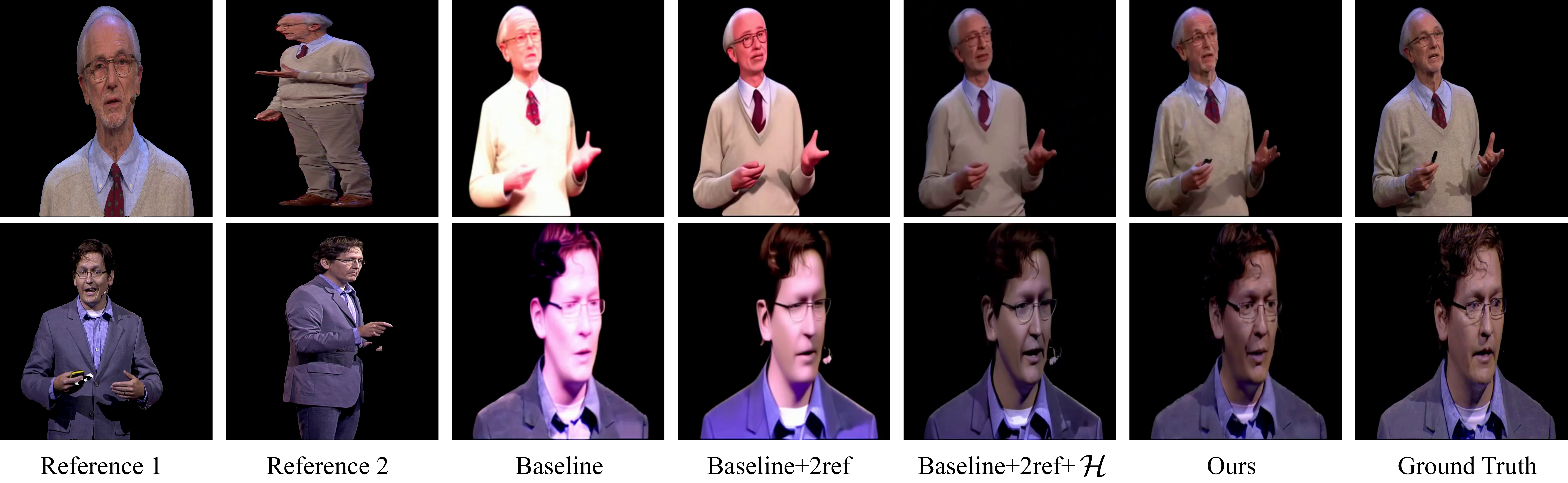}
\vspace{-20pt}
    \caption{The visualization of our ablation study. $\mathcal{H}$ is the pose correlation module. We can observe that with mulitple reference and pose correlation module, our method can obtain the best results.
    }
\vspace{-15pt}
    \label{fig:ablation}    
\end{figure*}

In this section, we conduct ablation studies to verify the effectiveness of each component of our framework. We remove the multiple reference mechanism and the reference selection strategy to establish a baseline. Based on this baseline, we perform ablation studies to evaluate the contributions of these components. Specifically, ``2ref'' indicates the use of two reference images, while $\mathcal{H}$ represents the pose correlation module.  The results of the ablation studies are presented in Table~\ref{tab:abla_ele} and Figure~\ref{fig:ablation}.

\begin{figure}[ht]
  \centering
    \includegraphics[width=1\linewidth]{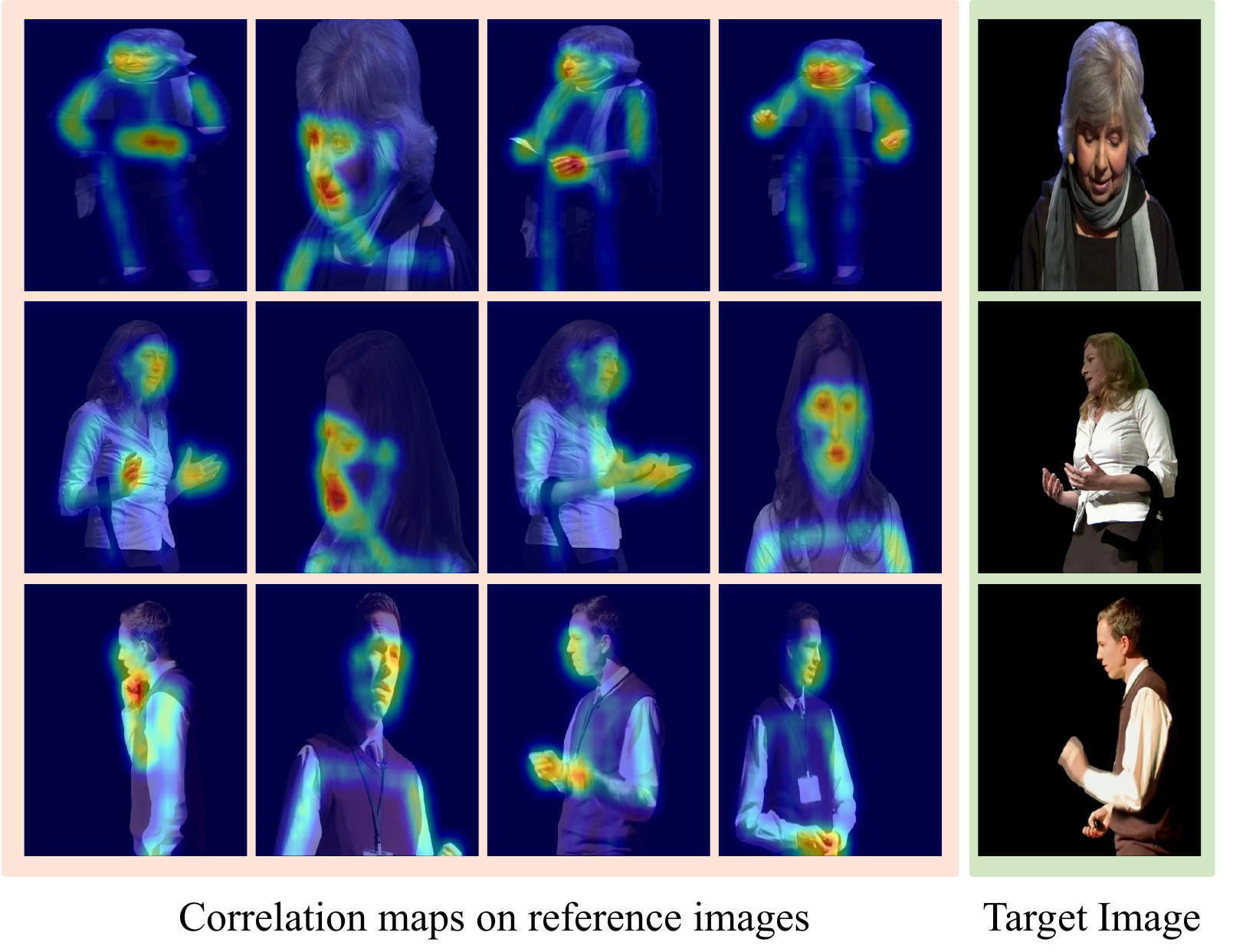}
\vspace{-20pt}
    \caption{Illustration of our learned correlation map. Our correlation map learned from the pose correlation module can indicate the informative region of the reference images.
    }
\vspace{-15pt}
    
    \label{fig:correlation_vis}    
\end{figure}
\noindent\textbf{Multiple Reference.}
In this work, we propose a multiple reference mechanism to synthesize free-viewpoint human animation videos. Compared to training with a single reference image, using multiple reference images provides more comprehensive visual patterns for the model, enhancing its ability to generalize and imagine, even when only a single reference image is used during inference.  To verify the effectiveness of using multiple reference images, we conduct experiments with different numbers of reference images during inference. As shown in Table~\ref{tab:comp_ted} and Table~\ref{tab:comp_dyhuman}, increasing the number of reference images leads to better results (Using two reference image in our model outperform using one reference by $0.31$ PSNR in Table~\ref{tab:comp_ted}). This improvement occurs because additional reference images provide richer visual patterns to synthesize videos from specific viewpoints.  Moreover, we perform an ablation study using two reference images in our baseline model. As shown in Figure~\ref{fig:ablation}, we can observe that with more one reference image, more details are included in the results (``Baseline'' vs. ``Baseline+2ref'') and the quantitative results show significant improvement, with FVD achieving the largest gain in Table~\ref{tab:abla_ele} ($9.82$ vs. $26.32$). The results, presented in Table~\ref{tab:comp_ted}, Table~\ref{tab:comp_dyhuman}, Table~\ref{tab:abla_ele} and Figure~\ref{fig:ablation}, further verify the effectiveness of using multiple reference images in free-viewpoint human animation video generation. 

\noindent\textbf{Correlation Map.}
Since multiple reference images can distract the attention of our model, we compute the correlation map between the target pose and each reference pose to enable the model to focus on the salient regions of the input reference images. As shown in Figure~\ref{fig:correlation_vis}, the correlation map highlights critical regions such as the head and hands of the person in the image. This indicates that our designed pose correlation module effectively enables the model to emphasize the informative regions of each reference image.
We present both qualitative and quantitative results in Table~\ref{tab:abla_ele} and Figure~\ref{fig:ablation}, which demonstrate improvements after deploying our pose correlation module. As show in Figure~\ref{fig:ablation}, we can observe that after deploy our pose correlation module, the overall quality of results obtain huge improvement. Because the model is able to percept the main region of reference images hinted by correlation attention map. We can also observe that quantitative improvement compared with the ablation ``baseline+2ref'' ($7.60$ vs. $9.82$ of FVD in Table~\ref{tab:abla_ele}) These results verify the effectiveness of our designed correlation map learning strategy.

\noindent\textbf{Reference Selection.}
As the number of references increases, the computational complexity can grow linearly. Therefore, we design a reference selection strategy that selects the most relevant reference features based on the correlation map. As shown in Table~\ref{tab:abla_ele} and Figure~\ref{fig:ablation}, the designed reference selection strategy enables our model to achieve slight but consistent improvements in visual quality, comparable to ``baseline+2ref+$\mathcal{R}$'' (PSNR: $32.49$ vs. $32.20$). These results verify that reference selection strategy can achieve comparable results while reducing inference time.

\vspace{-5pt}
\section{Conclusion}
\vspace{-5pt}
In this paper, we propose an adaptive reference selection diffusion network to tackle human animation under significant viewpoint changes. By incorporating a multiple reference mechanism, we provide comprehensive visual information for video synthesis. A correlation guider derives correlation maps between the target and reference poses, highlighting important areas. To reduce computational overhead, we introduce a reference selection strategy that chooses key reference features based on these maps. Additionally, we present a novel multi-shot TED video dataset (MSTed) to support further research in this domain. Our experiments show that multiple reference images enhance video quality, and our correlation map effectively identifies critical regions, leading to higher-fidelity results compared to state-of-the-art methods.

\vspace{-5pt}
\section{Limitation And Future Work}
\vspace{-5pt}
Our method relies on the quality and diversity of reference images, which may limit its performance in real-world applications. In future work, we plan to replace it with a more robust model, such as DiT, to improve results. Additionally, our MSTed dataset primarily consists of "talk" type data, and we aim to expand it by incorporating a variety of data sources for better generalization and robustness.

{
    \small
    \bibliographystyle{ieeenat_fullname}
    \bibliography{main}
}


\end{document}